# TOWARDS EFFICIENT VARIATIONAL AUTO-ENCODER USING WASSERSTEIN DISTANCE

*Zichuan Chen*     *Peng Liu\**

National University of Singapore    Singapore Management University

## ABSTRACT

VAE, or variational auto-encoder, compresses data into latent attributes and generates new data of different varieties. VAE with KL divergence loss has been considered an effective technique for data augmentation. In this paper, we propose using Wasserstein distance as a measure of distributional similarity for the latent attributes and show its superior theoretical lower bound ($ELBO_W$) compared with that of KL divergence ($ELBO_{KL}$) under mild conditions. Using multiple experiments, we demonstrate that the new loss function converges faster and generates better quality data to aid image classification tasks. We also propose implementing a dynamically changing hyper-parameter tuning schedule to avoid the potential overfitting of $ELBO_W$.

## 1 INTRODUCTION

Variational autoencoders (VAEs) come with different architectural variants. For example, (Aaron et al., 2017) and Razavi et al., 2019) use complex networks to reconstruct rich images distributions like shirts with figures and human faces with sunglasses. Compared with alternative deep generative models (i.e., generative adversarial networks, or GAN), VAE tend to suffer from blurriness and lack of sharpness in the reconstructed outputs. Though recent advances in VAE perform well in image deblurring, the complex architectures involved are more computationally intensive. Thus, computation efficiency is needed in autoencoder-based models.

This paper proposes a new loss function by replacing KL divergence with Wasserstein distance. Although the new loss function temporarily deviates from the theoretically motivated Evidence Lower Bound (ELBO), it has smaller empirical latent variances and converges faster with improved model performance in terms of the FID score. Given two-dimensional Wasserstein distance under Gaussian distribution, our approach bypasses the iterative computation of a high-dimensional Wasserstein distance thus can be computed in a single iteration as $ELBO_{KL}$.

We also note that our approach introduces a new hyper-parameter tuning schedule as motivated by the theoretical underpinnings of VAE. Specifically, comparing Wasserstein distance and KL divergence offers an inductive bias on tuning the hyper-parameter towards its theoretical optimum as the training proceeds.

## 2 RELATED WORK

VAE compresses the observed random variable $x$ into low-dimensional latent variable $z$ by an approximate posterior distribution $Q(z|x)$ as its encoder, which is then optimized together with the decoder network to recover $P(x)$. Originally proposed by (Kingma et al., at 2013), the vanilla VAE aims at maximizing $ELBO_{KL}$, which consists of marginal likelihood of the reconstructed data and a Kullback-Leibler (KL) divergence of the approximate and true posterior distribution on $z$:

$$ELBO_{KL} = E_{z \sim Q(z|x)}[\log(P(x|z))] = \int Q(z|x) \log(P(x|z)) \, dz - KL(Q(z|x) \,||\, P(z)) \quad (1)$$

where $KL(Q(z|x) \,||\, P(z)) = \int Q(z|x) \log \frac{Q(z|x)}{P(z)} dz$ measures the distance between $P(z)$, the true latent distribution (assumed to be standard normal) and approximate posterior $Q(z|x)$.

The Wasserstein distance $W_p(u, v)$, being a symmetric metric, is defined as an optimal transport from one distribution $u$ to the other $v$: $W_p(u, v) := \left( \inf_{\gamma \in \tau(u,v)} \int d(x,y)^p \, d\gamma(x,y) \right)^{1/p}$ where $\tau(u, v)$ denotes all joint distributions with the respective marginal distribution $u$ and $v$, $p$ is a constant larger than 1. Different from KL divergence, the Wasserstein distance would not explode with extreme value such as 0 or become meaningless when comparing two distributions without any overlap. Wasserstein

---

\* Correspondence author. Equal contribution.

Autoencoder (Tolstikhin et al, at 2017), or WAE, minimizes a form of Wasserstein Distance between model distribution and target data distribution. Depending on the penalty, there are two types of WAE: WAE-GAN using Jensen-Shannon divergence and WAE-MMD using penalty maximum mean discrepancy. Wasserstein-GAN (Arjovsky et al, at 2017) minimizes a robust and computationally efficient approximation of Wasserstein Distance in the loss function, presenting fast convergence and high-quality reconstructed image. Other examples include AdaGAN (Tolstikhin et al, at 2017), which adds a new component to the mixture model at every training step, and Sinkhorn Autoencoders (Patrini et al, at 2020), which minimizes the Wasserstein distance in the latent space using the Sinkhorn algorithm.

## 3 METHODOLOGY

We consider replacing the KL term in (1) by Wasserstein distance:

$$\text{ELBO}_W = \int Q(z|x) \log(P(x|z)) \, dz - W_p(Q(z|x) \,||\, P(z)) \quad (2)$$

Where we adopt $P(z) \sim N(0, I)$ and

$Q(z|x) \sim N((\mu_1, \mu_2, \ldots, \mu_m), \begin{pmatrix} \sigma_1^2 & \cdots & 0 \\ \vdots & \ddots & \vdots \\ 0 & \cdots & \sigma_m^2 \end{pmatrix})$, with the unknown mean vector $\{\mu_1, \mu_2, \ldots, \mu_m\}$ and diagonal variance vector $\{\sigma_1^2, \sigma_2^2, \ldots, \sigma_m^2\}$ to be estimated for independent latent variables in $z$. Since $P(z)$ is a multivariate normal, the 2nd Wasserstein distance can be reformulated as follows (Givens et al., at 1984):

$$W_2(Q(z|x) \,||\, P(z)) = ||\boldsymbol{\mu} - \mathbf{0}||_2^2 + Tr\left(\begin{pmatrix} \sigma_1^2 & \cdots & 0 \\ \vdots & \ddots & \vdots \\ 0 & \cdots & \sigma_m^2 \end{pmatrix} + \mathbf{I} - 2(\mathbf{I}^{\frac{1}{2}} \begin{pmatrix} \sigma_1^2 & \cdots & 0 \\ \vdots & \ddots & \vdots \\ 0 & \cdots & \sigma_m^2 \end{pmatrix}^{\frac{1}{2}} \mathbf{I}^{\frac{1}{2}})^{\frac{1}{2}}\right) \quad (3)$$

where $\boldsymbol{\mu} = (\mu_1, \mu_2, \ldots, \mu_m)$, $\mathbf{0}$ is a m-dimensional vector of 0, and $\mathbf{I}$ is a m-dimensional identity matrix. Computing the Wasserstein distance under such setting does not require solving an optimization problem and thus can be obtained in one shot.

Correspondingly, the total KL divergence can be written as:

$$KL(Q(z|x) \,||\, P(z)) = \frac{1}{2}(\prod_{i=1}^{m} -\log\sigma_i^2 + \sum_{i=1}^{m}(\mu_i^2 + \sigma_i^2) - m) \quad (4)$$

To compare $\text{ELBO}_W$ and $\text{ELBO}_{KL}$, we start with by their difference T:

$$T = W_2(Q(z|x) \,||\, P(z)) - KL(Q(z|x) \,||\, P(z)) = \log \prod_{i=1}^{m} \sigma_i^2 + \sum_i^m (\sigma_m - 2)^2 + \sum_i^m \mu_i^2 - m$$

Apparently, $T \leq 0 := \text{ELBO}_W \geq \text{ELBO}_{KL}$. Given sufficient iterations, the mean vector approaches 0 to match the prior, thus the value of T depends on the variances $\{\sigma_1, \sigma_2, \ldots, \sigma_m\}$:

1. When $\sigma_1, \sigma_{2,\ldots,}\sigma_m = 1$, $T = 0$, suggesting that $\text{ELBO}_W$ and $\text{ELBO}_{KL}$ are identical at this point.

2. When $\sigma_1, \sigma_{2,\ldots,}\sigma_m \leq 1$, it's easy to show $\frac{\partial T}{\partial \sigma_1}, \frac{\partial T}{\partial \sigma_2}, \ldots, \frac{\partial T}{\partial \sigma_m} \geq 0$, suggesting that T increases monotonically across $\{\sigma_1, \sigma_{2,\ldots,}\sigma_m\}$, jointly resulting in $T \leq 0$.

Therefore, when $\sigma_1, \sigma_{2,\ldots,}\sigma_m \leq 1$, $\text{ELBO}_W$ is closer to $\log(P(x))$ than $\text{ELBO}_{KL}$ in its approximation. Note that limiting the variance vector to fall into one of these two conditions can be achieved using a ReLU variant. To further aid the analysis, a 2-dimensional visualization is shown in Appendix A.

$\text{ELBO}_W$ and $\text{ELBO}_{KL}$ would be identical only when $Q(z|x)$ converges to $N(0, I)$. However, unlike $\text{ELBO}_{KL}$, $\text{ELBO}_W$ may not be a consistent estimator when the model overfits, and the training process may conclude with a higher $\text{ELBO}_W$ than $\log(P(x))$. This is illustrated in Appendix B. To enforce the consistency property of $\text{ELBO}_W$, it is necessary to place an inductive bias such that $T = 0$ upon model convergence. This is achieved by introducing an additional hyperparameter $\lambda$ that controls the weight of the Wasserstein distance term, forming a new objective function $\text{ELBO}_{W_\lambda}$ as follows.

$$\text{ELBO}_{W_\lambda} = \int Q(z|x) \log(P(x|z)) \, dz - \lambda W_2(Q(z|x) \,||\, P(z))$$

A large $\lambda$ imposes a high penalty on distributional dissimilarity. When $\lambda = 1$, $\text{ELBO}_{W_\lambda}$ fully recovers $\text{ELBO}_W$. The tuning schedule on $\lambda$ thus directly impacts the performance of $\text{ELBO}_{W_\lambda}$-based VAE.

Rather than following a random trial or an expensive grid search, dynamically adjusting $\lambda$ can help avoid overfitting by stay close to the inductive bias of $T = 0$. To this end, we propose a dynamic scheduling that involves an additional time series model on the sequence values of T:

$$\lambda_{t+1} = \lambda_t + \beta * c$$

$$\beta = \frac{1}{1 + e^{T_{t+1} - T_t}} - \frac{1 + \frac{T_{t+1}}{|T_{t+1}|}}{2}$$

where $c$ is a positive constant. Given $t$ iterations, $\lambda_{t+1}$ is updated through $\beta$, a parameter that depends on $T_{t+1}$ and $T_t$. In each iteration, we calculate the corresponding $T$ value and maintain a list $\{T_1, T_2, \ldots, T_{t-1}, T_t\}$. Based on this sequence data, a time series model such as ARIMA generates a predicted $T_{t+1}$, which is used to decide whether $\lambda_{t+1}$ should increase or decrease. Specifically, if $T_{t+1} \geq 0$, $\beta = \frac{1}{1+e^{T_{t+1}-T_t}} - 1$ is negative, thus $\lambda_{t+1}$ decreases, and a larger $T_{t+1} - T_t$ will incur a smaller $\lambda_{t+1}$. When $T_{t+1} \leq 0$, $\beta = \frac{1}{1+e^{T_{t+1}-T_t}}$ is positive, thus $\lambda_{t+1}$ increases. See an implementation of this approach in Appendix C.

## 4 EXPERIMENTS

In this section, we illustrate the effectiveness of data augmentation using $ELBO_W$ in image classification tasks based on MNIST and Fashion-MNIST datasets. We use FID score to measure the quality of the generated images, where a lower FID score corresponds to higher similarity between the original image and the generated image. For this experiment, we use a 3-layer MLP architecture for both encoder and decoder and allow $\lambda$ to vary uniformly between 0 and 20. We select the optimal $\lambda$ across 30 experiments with different random seeds. For each trained model, we set the number of epochs to 15, the batch size to 100, and transform each image into a standard 28 by 28 matrix. Note that we monitor the value of T and exclude results when the model overfits or underfits. Table 1 contains the mean FID scores and the standard deviation across 30 runs for both datasets. The lower FID score using $ELBO_W$ shows that our proposed model could generate higher-quality images compared with $ELBO_{KL}$ and WAE-MMD in MNIST-based task. (Codes are written under PyTorch 1.9.0, Paszke et al., at 2019) We also compare with WAE-MMD (from Tolstikhin et al, at 2017) in the MNIST experiment, with mean values of FID scores 56.6 and standard deviation 3.5.

| MNIST | | Fashion_MNIST | |
|---|---|---|---|
| $ELBO_W$ | $ELBO_{KL}$ | $ELBO_W$ | $ELBO_{KL}$ |
| **23.1 (2.5)** | 81.7 (4.3) | **108.2 (7.1)** | 230 (10.4) |

Table 1-Mean and Standard variance of FID scores for $ELBO_W$ and $ELBO_{KL}$ in 30 runs

To better detect the marginal value of data augmentation, the generated images are merged into the training set to train the image classifier. Specifically, we randomly selected 10,000 generated images based on the respective best trained model using $ELBO_W$ and $ELBO_{KL}$, and merged with the original dataset of different sizes, namely 10,000, 20,000 and 60,000. A separate 3-layer CNN classification model is trained based on multiple datasets, with another 3,000 images selected as the test dataset to measure the predictive accuracy. The results, as shown in table 2, suggest that $ELBO_W$-based VAE model better aids the classification task, especially when the original data size is relatively small and under-represented.

| Training data size | Original data (baseline) | Original + $ELBO_{KL}$ augmentation | Original + $ELBO_W$ augmentation |
|---|---|---|---|
| 10000 | 30.20% | 65.40% | **72.90%** |
| 20000 | 65.10% | 76.40% | **77.80%** |
| 60000 | 86.70% | **87.40%** | 87.20% |

Table 2-Prediction accuracy using an additional 10000 augmented images reconstructed from $ELBO_{KL}$ and $ELBO_W$ based VAE models

Moreover, the $ELBO_W$ based model converges faster than $ELBO_{KL}$ according to the loss profile across 30 runs at 1000th, 2000th and 5000th iteration as shown in Table 3.

| Iterations | $ELBO_{KL}$ based VAE Loss | $ELBO_W$ based VAE Loss |
|---|---|---|
| 100 | 22458.13 | 16491.7 |
| 1000 | 18752.58 | 8239.5 |
| 2000 | 14331.29 | **7768.86** |
| 5000 | **12674.95** | 7662.49 |

Table 3-Average loss at 1000th, 2000th, 5000th iteration from 30 runs.

## 5 CONCLUSION

This paper proposes a new objective function for variational auto-encoder, a widely used image generation architecture in data augmentation. By replacing KL divergence with Wasserstein distance to form a new variational lower bound, we show that $ELBO_W$ has a better theoretical approximation and more efficient convergence than $ELBO_{KL}$ under mild conditions. The conditions are easily satisfied using an adjusted ReLU activation function, making it convenient for practical implementation. We also propose a new tuning scheduling for the hyperparameter based on theoretically motivated inductive bias. Experiment results show that $ELBO_W$-based VAE could not only generate artificial images of higher quality, but also better improves classification accuracy compared with $ELBO_{KL}$-based model, especially when the original training data is small.

**Appendix A**

We limit the analysis to a two-dimensional encoder output for the mean and variance vectors, thus we have:

$T \leq 0 := \log \sigma_1^2 \sigma_2^2 + (\sigma_1 - 2)^2 + (\sigma_2 - 2)^2 + \mu_1^2 + \mu_2^2 \leq 2$

Below Figure 1 shows T with different variance values $\sigma_1$ and $\sigma_2$. Given enough iterations, the mean values will approach $0^1$, and we can derive the monotonicity and upper bound of $T$ as a function of $\sigma_1$ and $\sigma_2$: 1. When $\sigma_1, \sigma_2 = 1$, $T = 0$, $ELBO_W$ and $ELBO_{KL}$ have no difference. 2. When $\sigma_1, \sigma_2 \leq 1$, $\frac{\partial T}{\partial \sigma_1}, \frac{\partial T}{\partial \sigma_2} \geq 0$, T increases as $\sigma_1, \sigma_2$ increase, therefore $T \leq 0$.

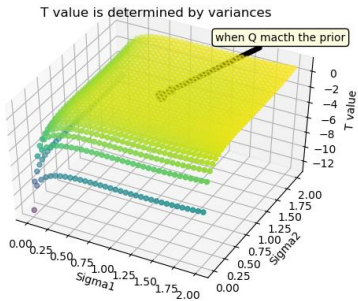

Figure 1-T value, which is distance between $ELBO_W$ and $ELBO_{KL}$, varies by latent variance

**Appendix B**

In perfect condition, $ELBO_W$ converges to the same point as $ELBO_{KL}$: $ELBO_{KL} = ELBO_W = -\int P(z) \log(P(x|z)) dz^2$. However, in general, $ELBO_W$ is always larger than $ELBO_{KL}$ when constraining the variances using ReLU variant. A potential problem is that $ELBO_W$ may exceed the theoretical upper bound in $ELBO_{KL}$ case, $\log(P(x))$. In other words, we expect condition 1 but as epoch goes, we may reach to condition 2, as shown below:

Condition 1: $ELBO_{KL} \leq ELBO_W \leq \log(P(x))$
Condition 2: $ELBO_{KL} \leq \log(P(x)) \leq ELBO_W$

If Condition 2 happens, $ELBO_W$ may approach infinity due to the term $\log \prod_{i=1}^{m} \sigma_i^2$, and the model may overfit. To avoid Condition 2, a naïve solution is to set T as a regularization term.

In Figure 2, we present Conditions 1 and 2 in training and its corresponding result. In avoid of condition 2, a regularization term is imperative and useful.

---

[1] Mean values are forced to 0 by minimizing ELBO/ELBO_W given sufficient iterations

[2] In computation we actually sample from the distribution of z (called reparameterization trick) so the equation is satisfied given samples z (dimensions of encoder output layer) large.

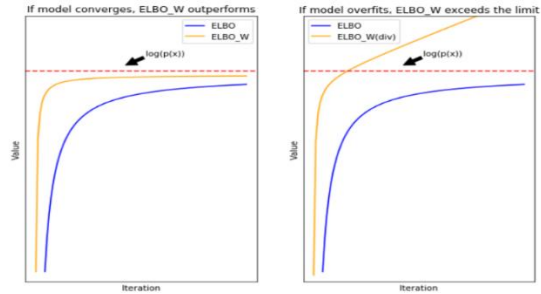

Figure 2- $ELBO_W$ based VAE is unlikely to overfit with regularization term in loss

**Appendix C**

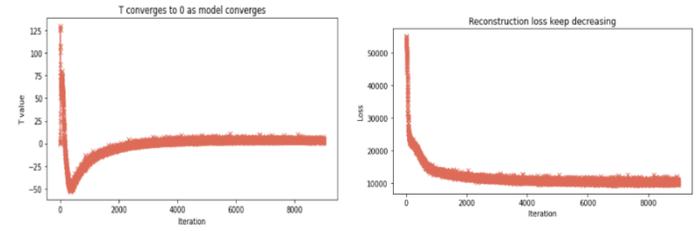

Figure 3,4-With regularization term, T value returns to 0 when model, while reconstruction loss still decreases

Figure 3 and 4 shows profile of the T value and the reconstruction loss with a dynamically scheduled $\lambda$. The model firstly focuses on minimizing the cross-entropy term $-\int Q(z|x) \log(P(x|z)) dz$ since $\lambda$ is small at initial iterations, but gradually upweights the distance term $W(Q(z|x) || P(z))$ as iteration continues. Finally, the model is expected to converge at a point when T = 0. In our experiments, initial value of $\lambda$ is set to 1, the constant $c$ is 0.05, and all other hyperparameters follow the same setting as in section 4. While T converges to 0, making $ELBO_W$ equal to $ELBO_{KL}$, the reconstruction loss keeps decreasing as training proceeds. Figure 5 shows two sample images reconstructed using $ELBO_W$ based VAE and $ELBO_{KL}$ based VAE respectively. The left image, as reconstructed by $ELBO_W$ based VAE, demonstrates an obvious visual clarity and sharpness than the left image under the same training budget.

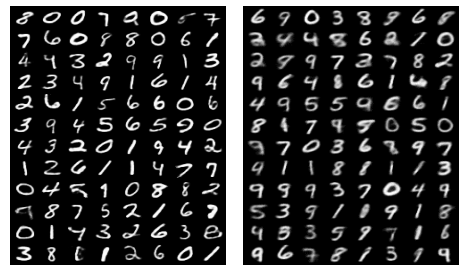

Figure 5- Left: Reconstructed by $ELBO_W$ based VAE Right: Reconstructed by $ELBO_{KL}$ based VAE